\documentclass[twoside,11pt]{article}

\usepackage{blindtext}

\usepackage{graphicx}
\usepackage{subcaption}
\usepackage{jmlr2e}
\usepackage{lastpage}
\jmlrheading{26}{2025}{1-\pageref{LastPage}}{8/24; Revised
11/25}{11/25}{24-1353}{Sarah Segel, Helena Graf, Edward Bergman, Kristina Thieme, Marcel Wever, Alexander Tornede, Frank Hutter, Marius Lindauer}

\ShortHeadings{\Deepcave: A Visualization and Analysis Tool for Automated Machine Learning}{Segel, Graf, Bergman, Thieme, Wever, Tornede, Hutter, and Lindauer}
\firstpageno{1}

\begin{document}

\newcommand{\HPO}{HPO\xspace}
\newcommand{\ML}{ML\xspace}
\newcommand{\AutoML}{AutoML\xspace}
\newcommand{\Deepcave}{DeepCAVE\xspace}

\title{\Deepcave: A Visualization and Analysis Tool for Automated Machine Learning}

\author{\name Sarah Segel$^1$ \email s.segel@ai.uni-hannover.de
       \AND
       \name Helena Graf$^1$ \email h.graf@ai.uni-hannover.de
       \AND
       \name Edward Bergman$^2$ \email bergmane@cs.uni-freiburg.de
       \AND
       \name Kristina Thieme$^1$ \email kristina.thieme@stud.uni-hannover.de
       \AND
       \name Marcel Wever$^{1,4}$ \email m.wever@ai.uni-hannover.de
       \AND
       \name Alexander Tornede$^1$ \email a.tornede@ai.uni-hannover.de
       \AND
       \name Frank Hutter$^{2,3}$ \email fh@cs.uni-freiburg.de
       \AND
       \name Marius Lindauer$^{1,4}$ \email m.lindauer@ai.uni-hannover.de
       \\
       \addr $^1$Leibniz University Hannover, 
       \addr $^2$University of Freiburg,
       \addr $^3$ELLIS Institute T{\"u}bingen,
       \addr $^4$L3S Research Center
       }

\editor{Zeyi Wen}

\maketitle

\begin{abstract}
Hyperparameter optimization (\HPO), as a central paradigm of AutoML, is crucial for leveraging the full potential of machine learning (ML) models; yet its complexity poses challenges in understanding and debugging the optimization process. 
We present \Deepcave, a tool for interactive visualization and analysis, providing insights into \HPO. 
Through an interactive dashboard, researchers, data scientists, and ML engineers can explore various aspects of the \HPO process and identify issues, untouched potentials, and new insights about the ML model being tuned. 
By empowering users with actionable insights, \Deepcave contributes to the interpretability of \HPO and ML on a design level and aims to foster the development of more robust and efficient methodologies in the future.
\end{abstract}

\begin{keywords}
  Automated machine learning, hyperparameter optimization, interpretable machine learning, human-centered machine learning, visualization
\end{keywords}

\section{Introduction}

Modern methods for hyperparameter optimization (\HPO) encompass a diverse range of approaches beyond grid and random search~\citep{turner-neuripscomp21a,bischl-dmkd23a}.
However, their lack of insight and transparency can lead to a lack of trust by users~\citep{drozdal-iui20a} and make debugging such methods and their application difficult.
We introduce \Deepcave, a visualization and analysis tool for \HPO, aiding in understanding and debugging the application of \HPO.
At present, \HPO is our main priority, wherein metadata in the format of configurations and performance values is generated in an iterative process, as many automated machine learning (\AutoML) packages do, such as Auto-Sklearn~\citep{feurer-jmlr22a}, TPOT~\citep{olson-automl19b}, AutoPyTorch~\citep{zimmer-tpami21a} or Auto-Keras~\citep{jin-sigkdd19a}.
Similarly, leveraging \Deepcave for neural architecture search (NAS)~\citep{elsken-arxiv22a} or combined algorithm selection and hyperparameter optimization~\citep{thornton-kdd13a} is possible by encoding architectural decisions, algorithmic choices, or machine learning pipelines as part of the configuration.
Advances in machine learning and \HPO, including multi-objective or multi-fidelity methods, challenge interpretation and analysis tools.
\Deepcave addresses this by offering support for multiple objectives and fidelities, with specific visualizations tailored to the different scenarios.
This tightly interacts with the vision of a more human-centered approach of AutoML~\citep{lindauer-icml24a}, in which users can learn from \HPO and ultimately even integrate their expertise into the system~\citep{hvarfner-iclr22, mallik-neurips23a}.
Moreover, to facilitate a reproducible and transparent research process, users can export visualizations to incorporate into their research papers, thereby providing insight into their \HPO process and optimized hyperparameters.

\section{Features and Usage}

\begin{figure}
    \centering
    \includegraphics[width=.65\textwidth]{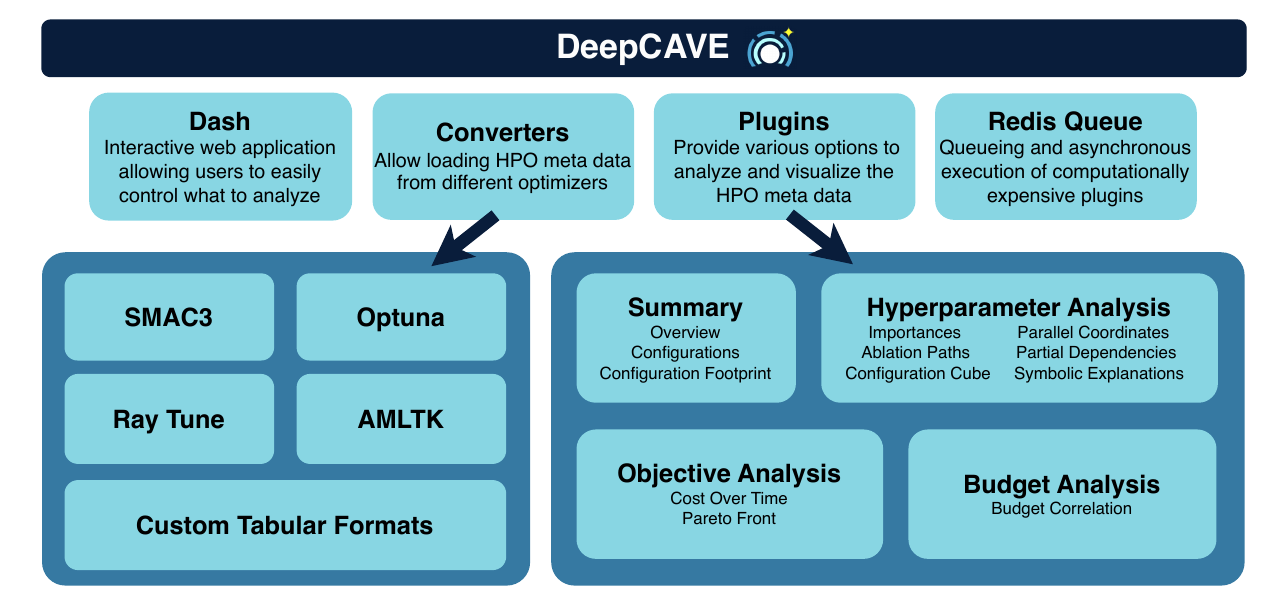}
    \caption{Overview of \Deepcave's components and their functionalities.}
    \label{fig:structure}
\end{figure}

\Deepcave's architecture relies upon various components, as illustrated in \autoref{fig:structure}. 
The tool is exclusively written in Python, given its prevalence as the predominant programming language in machine learning~\citep{raschka-information20}.
Built on the Dash framework~\citep{plotly-15a}, it offers users a fully interactive environment within a web browser.

\subsection{Converters}

\Deepcave utilizes run objects as a basic unit for data interpretation, where a run can be considered an \HPO process consisting of a list of trials, each representing a hyperparameter configuration with associated objective value, budget, and seed. 
The interface allows for selection and grouping of runs, streamlining the analysis of many \HPO runs at once. 
Converters are used to access the optimizer data stored in the file system and transfer them into a run object. 
They keep track of both finished and running optimization processes that consistently write new results to disk. 
At the time of writing, \Deepcave supports loading runs created by the \HPO packages SMAC3~\citep{lindauer-jmlr22a}, Ray Tune~\citep{liaw-automl18a}, and Optuna~\citep{akiba-ickddm19a}, as well as, the \AutoML framework AMLTK~\citep{bergman2023amltk}.
Furthermore, to allow loading runs from arbitrary optimizers, \Deepcave offers a generic tabular input format, as well as a programmatic interface.

\subsection{Analysis Plugins}

\Deepcave's strength lies in its modular plugin structure, providing diverse insights into the \HPO run. 
Plugins allow analyzing overall runs, objectives, hyperparameters, and budgets through texts, tables, and Plotly~\citep{plotly-15a} visualizations. 
Dynamic filtering enables the interactive selection of certain aspects, such as specific objectives or budgets.
For computationally intensive plugins, Redis Queue~\citep{redis-queue} is leveraged for efficient queueing and asynchronous execution, ensuring a responsive user experience.

\begin{figure}
    \centering
    \begin{subfigure}[b]{0.35\textwidth}
        \centering
        \includegraphics[width=\textwidth]{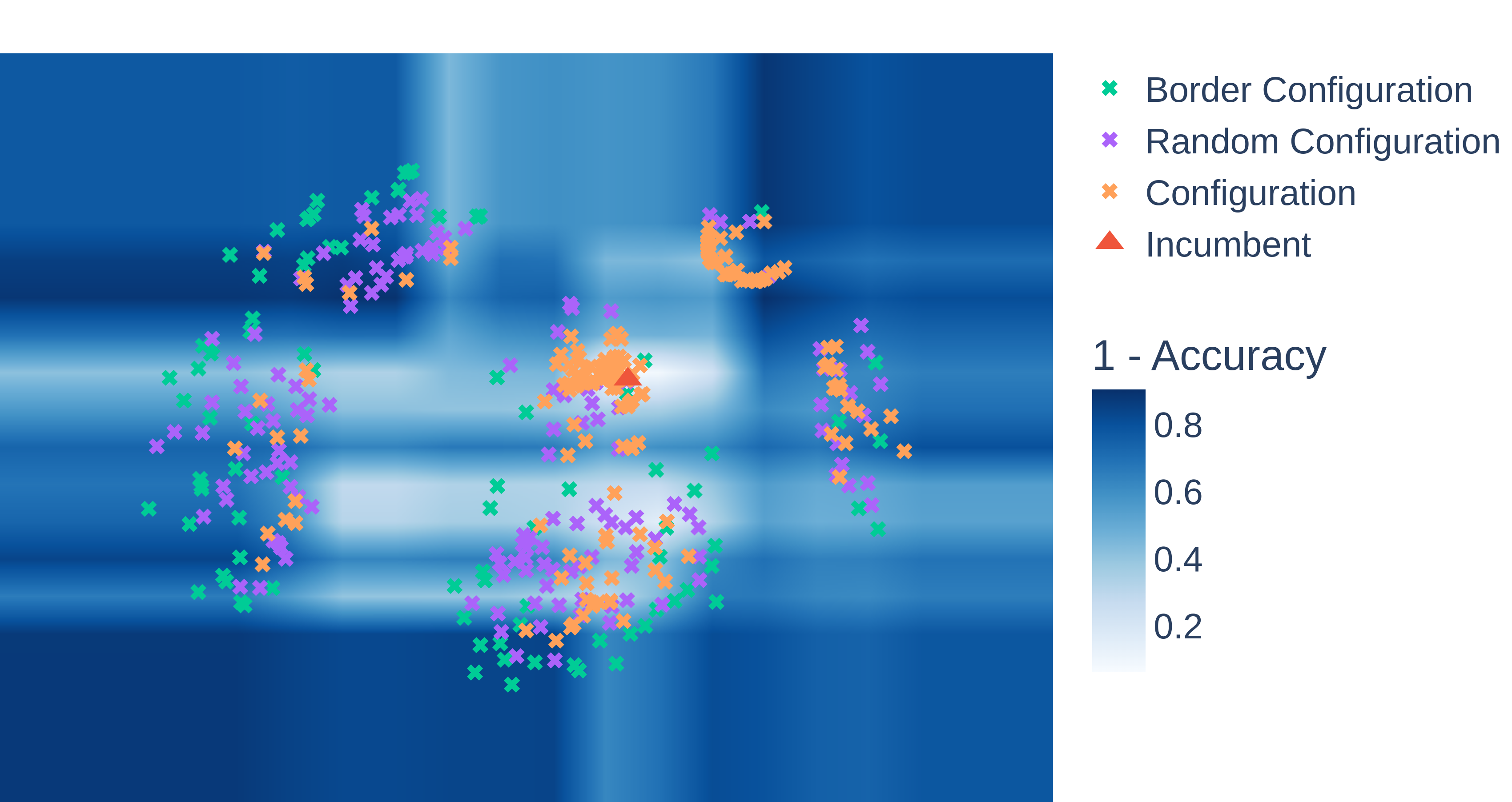}
        \vspace{0.09\baselineskip}
        \caption{Configuration Footprint.}
        \label{fig:footprint}
    \end{subfigure}
    \begin{subfigure}[b]{0.32\textwidth}
        \centering
        \includegraphics[width=\textwidth]{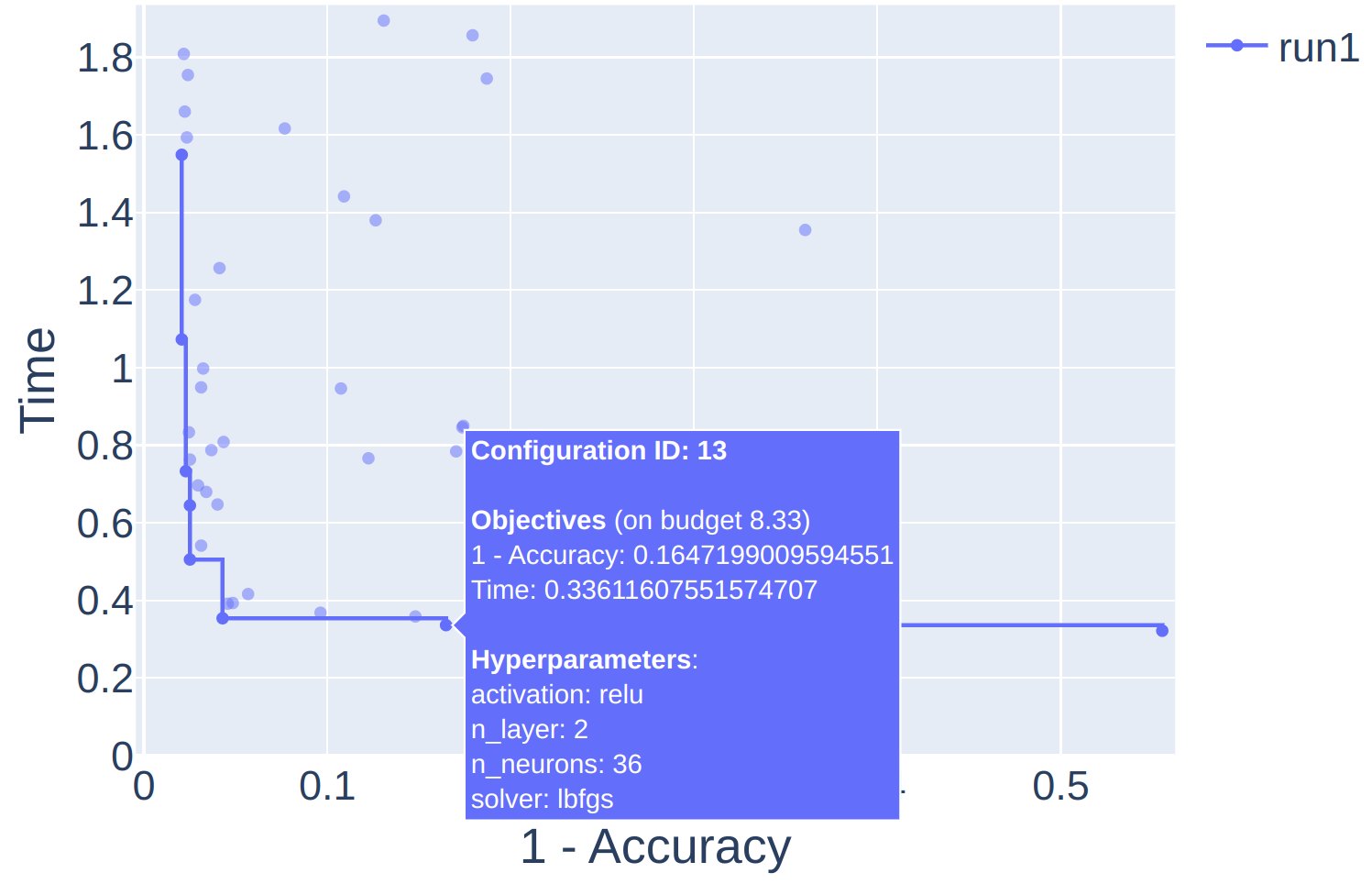}
        \vspace{-0.92\baselineskip}
        \caption{Pareto Front.}
        \label{fig:pareto-front}
    \end{subfigure}
    \begin{subfigure}[b]{0.31\textwidth}
        \centering
        \includegraphics[width=\textwidth]{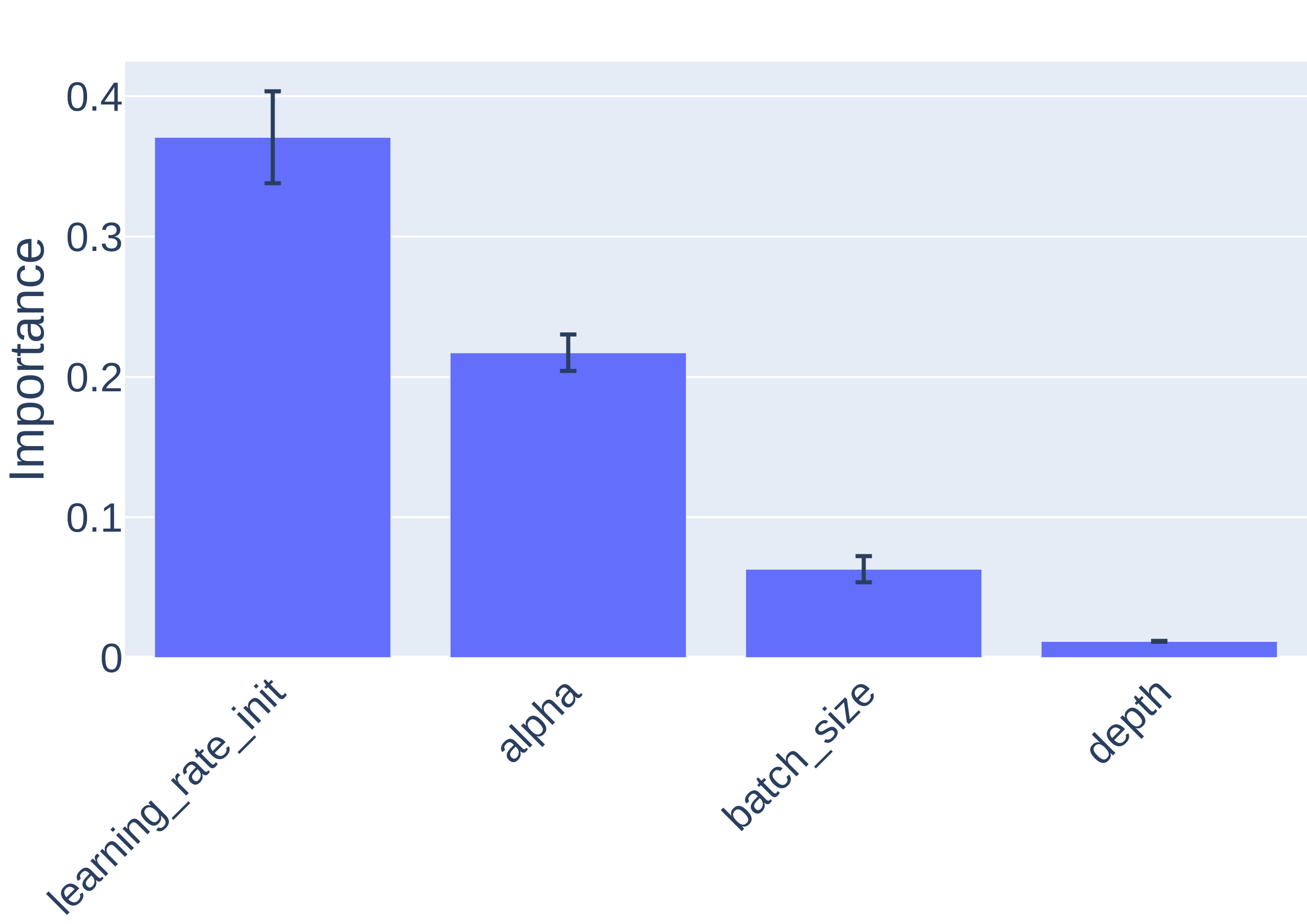}
        \vspace{-0.91\baselineskip}
        \caption{Importances.}
        \label{fig:importances}
    \end{subfigure}
    \caption{Examples of plots produced via \Deepcave's plugins. A mouseover allows obtaining additional information, and clicking on single configurations in the plot opens the Configurations plugin, providing details regarding that configuration.}
    \label{fig:whole}
\end{figure}

\textbf{Overall Optimization Analysis} The \textit{Overview plugin} offers a holistic view, showing optimizer choice, configuration space, and trial status.
Detailed analysis of individual configurations can be performed via the \textit{Configurations plugin}.
Employing multi-dimensional scaling~\citep{kruskal-psychometrika64a, kruskal-psychometrika64b} for dimensionality reduction, the \textit{Configuration Footprint plugin} visually represents the optimizer's coverage of the configuration space~\citep{biedenkapp-lion18a}, as shown in~\autoref{fig:footprint}.
This can help to identify limited exploration, suggesting the optimizer might be stuck in a local optimum.

\textbf{Objective Analysis} The \textit{Cost Over Time plugin} visualizes optimization convergence, either over time or the number of trials evaluated, allowing to compare different optimization strategies such as the optimizer chosen, search space design, or other meta-level decisions. 
The \textit{Pareto Front plugin} visually represents configurations that achieve different tradeoffs between conflicting objectives, such as runtime versus final loss, as shown in~\autoref{fig:pareto-front}.

\textbf{Hyperparameter Analysis} As the relationship between hyperparameter configurations and objective values constitutes a crucial aspect of \HPO, \Deepcave provides a variety of plugins in this regard.
\textit{Parallel Coordinates} plots visualize hyperparameter configurations as lines, connecting their hyperparameter values and corresponding final scores~\citep{golovin-kdd17a}. 
Additionally, \textit{Partial Dependence} plots~\citep{friedman-as01a, moosbauer-neurips21a} offer insights into the average marginal effect of single hyperparameters on the objective value. 
Moreover, \textit{Symbolic Explanations}~\citep{segel-automl23a} provide explicit formulas capturing the relationship between hyperparameter and objective values.
To understand the impact of individual hyperparameters on the objective, the \textit{Importances plugin} can be used.
With fANOVA~\citep{hutter-icml14a}, see~\autoref{fig:importances}, and local (hyper)parameter importance (LPI)~\citep{biedenkapp-lion18a}, users can explore both global and local importance.
\textit{Ablation Paths}~\citep{biedenkapp-aaai17a} provide insights into the impact of sequentially adjusting individual hyperparameter values towards the incumbent configuration. In practice, \Deepcave has been used to analyze optimization runs with up to 39 hyperparameters, confirming its scalability for high-dimensional settings.

\textbf{Budget Analysis} Effective use of multiple fidelities crucially relies on a consistent rank or performance correlation to effectively allocate more resources to better configurations.
The \textit{Budget Correlation plugin} directly visualizes this correlation between fidelities, allowing users to assess the utility of their multiple fidelities.

Although \Deepcave provides diverse analysis plugins and visualizations, their customization and composition are currently limited, outlining interesting future work.

\section{Existing Tools for Explainable \AutoML}
With the rising interest in explainability and insight into optimization processes, a number of tools aiming to fill this gap have emerged. 
Among these, XAutoML~\citep{zoller-acm23a} focuses on \AutoML and interpretability. 
Similar to \Deepcave, it provides a number of different interactive visualizations.
Notably, it allows users to extend its capabilities through custom ``cards'', similar to plugins in \Deepcave, and supports a range of common (Auto)ML libraries. 
In contrast to our tool, it lacks support for (parallel) background computation and analysis of running \HPO processes.
Further, only one \HPO run is considered at a time, making comparisons between runs cumbersome. 
Beyond XAutoML, platforms like RapidMiner~\citep{altair}, Google Vizier~\citep{golovin-kdd17a}, and ATMseer~\citep{wang-chi18a} cater to specific facets of \AutoML processes, such as algorithm selection, experiment tracking, and pipeline analysis, albeit often with limitations in terms of interpretability at the \HPO level or support for run comparisons. 
Similarly, tools like Optuna~\citep{akiba-ickddm19a}, IOHanalyzer~\citep{wang-acm22a}, TensorBoard~\citep{abadi-15a}, WandB~\citep{biewald-wandb20a}, MLFlow~\citep{chen-deem20a}, MLJar~\citep{plonska-21a}, SigOpt~\citep{clark-19a}, IBM AutoAI~\citep{wang-iui20ab}, Pipeline Profiler~\citep{ono-ieee21a}, iraceplot~\citep{lopezibanez-github25}, and Boxer~\citep{gleicher-cgf20a} contribute valuable insights through various forms of run analysis, visualization, and reporting. 
However, many of these tools depend on specific formats, lack extensibility, or fail to support comparisons across multiple \HPO runs.
\Deepcave distinguishes itself in this field with its focus on \HPO, interpretability, and interactivity. 
Its unique proposition lies in its ability to facilitate detailed comparisons across different \HPO runs.
Moreover, its extensive format compatibility, efficient computational parallelization, and browser-based accessibility mitigate major usability challenges found in comparable tools.

\section{Conclusion}
\Deepcave is a novel tool for interactively analyzing and explaining \HPO runs. 
It allows performing comprehensive analyses of \HPO runs concerning their hyperparameters, objectives, and budgets via an interactive dashboard running in the web browser.
\Deepcave is accessible under Apache License Version 2.0\footnote{\url{https://www.apache.org/licenses}} and can be installed via the instructions in the GitHub repository \url{https://github.com/automl/DeepCAVE}.


\acks{Sarah Segel acknowledges financial support by the Federal Ministry for Economic Affairs and Energy of Germany in the project CoyPu under Grant No. 01MK21007L. Helena Graf, Alexander Tornede, Marcel Wever, and Marius Lindauer gratefully acknowledge funding by the European Union (ERC, ``ixAutoML'', grant no.101041029). Frank Hutter acknowledges funding by the European Union (ERC Consolidator Grant DeepLearning 2.0, grant no.~101045765).
Edward Bergman was partially supported by TAILOR, a project funded by EU Horizon 2020
research and innovation programme under GA No 95221. Views and opinions expressed are however those of the author(s) only and do not necessarily reflect those of the European Union or the European Research Council Executive Agency. Neither the European Union nor the granting authority can be held responsible for them.

\begin{center}
    \includegraphics[height=2cm]{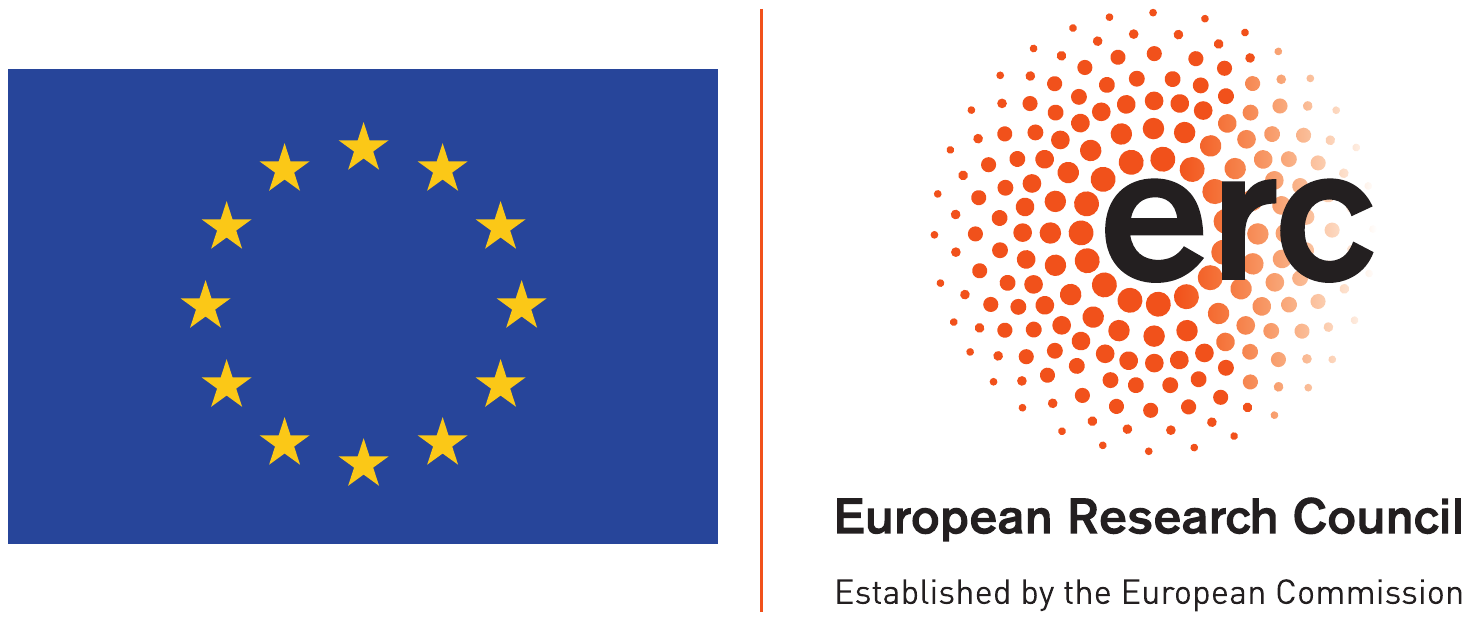}
\end{center}
}



\bibliography{strings, lib, additional, proc}

@inproceedings{lindauer-icml24a,
  author       = {M. Lindauer and
                  F. Karl and
                  A. Klier and
                  J. Moosbauer and
                  A. Tornede and
                  A. Mueller and
                  F. Hutter and
                  M. Feurer and
                  B. Bischl},
  title        = {Position: {A} Call to Action for a Human-Centered {AutoML} Paradigm},
  booktitle      = {Proc. of {ICML}'24},
  year         = {2024},
}

@inproceedings{liaw-automl18a,
  author       = {R. Liaw and
                  E. Liang and
                  R. Nishihara and
                  P. Moritz and
                  J. Gonzalez and
                  I. Stoica},
  title        = {Tune: {A} Research Platform for Distributed Model Selection and Training},
  crossref={automl18}
}

@article{bergman2023amltk,
  title={{AMLTK}: A Modular {A}uto{ML} Toolkit in {P}ython},
  author={E. Bergman and M. Feurer and A. Bahram and A. R. Balef and L. Purucker and S. Segel and M. Lindauer and F. Hutter and K. Eggensperger},
  journal={Journal of Open Source Software},
  year={2024},
  publisher = {The Open Journal},
  volume = {9},
  number = {100},
  pages = {6367}, 
}

@article{kruskal-psychometrika64a,
  title={Multidimensional scaling by optimizing goodness of fit to a nonmetric hypothesis},
  author={J. Kruskal},
  journal={Psychometrika},
  year={1964},
  volume={29},
  pages={1-27},
}

@article{kruskal-psychometrika64b,
  title={Nonmetric multidimensional scaling: A numerical method},
  author={J. Kruskal},
  journal={Psychometrika},
  year={1964},
  volume={29},
  pages={115–129},
}

@article{raschka-information20,
AUTHOR = {S. Raschka and J. Patterson and C. Nolet},
TITLE = {Machine Learning in {P}ython: Main Developments and Technology Trends in Data Science, Machine Learning, and Artificial Intelligence},
JOURNAL = {Information},
VOLUME = {11},
YEAR = {2020},
NUMBER = {4},
ARTICLE-NUMBER = {193},
}

@misc{plotly-15a, 
    author = {{Plotly Technologies Inc.}}, 
    title = {Collaborative data science}, 
    publisher = {Plotly Technologies Inc.}, 
    year = {2015}, 
    url = {https://plot.ly},
}

@article{zoller-acm23a,
author = {M. Z{\"{o}}ller and W. Titov and T. Schlegel and M. F. Huber},
title = {{XA}uto{ML}: {A} Visual Analytics Tool for Understanding and Validating Automated Machine Learning},
year = {2023},
publisher = {ACM Press},
volume = {13},
number = {4},
journal = {ACM Transactions on Interactive Intelligent Systems},
articleno = {28},
}

@misc{altair,
    author={{Altair Engineering Inc.}},
    title={Altair {R}apidMiner Data Analytics and {AI} Platform},
    url={https://altair.com/altair-rapidminer},
    year={2024},
}

@inproceedings{wang-chi18a,
	title = {{ATMSeer}: Increasing Transparency and Controllability in Automated Machine Learning},
	author = {Q. Wang and Y. Ming and Z. Jin and Q. Shen and D. Liu and M. J. Smith and K. Veeramachaneni and H. Qu},
	booktitle = {Proceedings of the 2019 CHI Conference on Human Factors in Computing Systems},
	year = {2018}
}

@inproceedings{akiba-ickddm19a,
	title = {Optuna: A Next-generation Hyperparameter Optimization Framework},
	author = {T. Akiba and S. Sano and T. Yanase and T. Ohta and M. Koyama},
	booktitle = {Proc. of KDD'19},
	year = {2019}
}

@article{wang-acm22a,
	title = {{IOHanalyzer}: Detailed Performance Analyses for Iterative Optimization Heuristics},
	author = {H. Wang and D. Vermetten and F. Ye and C. Doerr and T. B\"{a}ck},
	journal = {ACM Transactions on Evolutionary Learning and Optimization},
	volume = {2},
	number = {1},
	year = {2022}
}

@misc{biewald-wandb20a,
	title = {Experiment Tracking With {W}eights and {B}iases},
	author = {L. Biewald},
	year = {2020},
        url={https://www.wandb.com/},
    }

@misc{abadi-15a,
	title = {{TensorFlow}: Large-scale Machine Learning on Heterogeneous Systems},
	author = {M. Abadi and A. Agarwal and P. Barham and E. Brevdo and Z. Chen and C. Citro and G. Corrado and A. Davis and J. Dean and M. Devin and S. Ghemawat and I. Goodfellow and A. Harp and G. Irving and M. Isard and Y. Jia and R. Jozefowicz and L. Kaiser and M. Kudlur and J. Levenberg and D. Man\'{e} and R. Monga and S. Moore and D. Murray and C. Olah and M. Schuster and J. Shlens and B. Steiner and I. Sutskever and K. Talwar and P. Tucker and V. Vanhoucke and V. Vasudevan and F. Vi\'{e}gas and O. Vinyals and P. Warden and M. Wattenberg and M. Wicke and Y. Yu and X. Zheng},
	year = {2015},
        url={https://www.tensorflow.org},
}

@inproceedings{chen-deem20a,
	title = {Developments in {MLflow}: {A} System to Accelerate the Machine Learning Lifecycle},
	author = {A. Chen and A. Chow and A. Davidson and A. DCunha and A. Ghodsi and S. Hong and A. Konwinski and C. Mewald and S. Murching and T. Nykodym and P. Ogilvie and M. Parkhe and A. Singh and F. Xie and M. Zaharia and R. Zang and J. Zheng and C. Zumar},
	booktitle = {Proc. of International Workshop on Data Management for End-to-End Machine Learning},
	year = {2020}
}

@misc{plonska-21a,
	title = {{MLJAR}: State-of-the-art Automated Machine Learning Framework for Tabular Data. Version 0.10.3},
	author = {A. P\l{}o\'{n}ska and P. P\l{}o\'{n}ski},
	year = {2021},
        url={https://mljar.com},
}

@misc{clark-19a,
	title = {Sigopt Web Page},
	author = {S. Clark and P. Hayes},
	year = {2019},
        url={https://sigopt.org},
    }

@article{ono-ieee21a,
	title = {{PipelineProfiler}: A Visual Analytics Tool for the Exploration of {AutoML} Pipelines},
	author = {J. Ono and S. Castelo and R. Lopez and E. Bertini and J. Freire and C. Silva},
	journal = {IEEE Transactions on Visualization and Computer Graphics},
	volume = {27},
	number = {2},
	pages = {390-400},
	year = {2021}
}

@article{gleicher-cgf20a,
	title = {Boxer: Interactive Comparison of Classifier Results},
	author = {M. Gleicher and A. Barve and X. Yu and F. Heimerl},
	journal = {Computer Graphics Forum},
	volume = {39},
	number = {3},
	year = {2020}
}

@misc{redis-queue,
    author = {{Redis Inc.}},
    title = {Redis {Q}ueue},
    url={https://github.com/rq/rq},
    year={2024}
}

@inproceedings{olson-automl19b,
  title        = {{TPOT}: A Tree-Based Pipeline Optimization Tool for Automating Machine Learning},
  author       = {R. Olson and J. Moore},
  pages        = {151--160},
  booktitle    = {Automated Machine Learning: Methods, Systems, Challenges},
  year         = 2019,
  publisher    = {Springer},
  note         = {Available for free at \url{http://automl.org/book}},
  editor       = {F. Hutter and L. Kotthoff and J. Vanschoren},
}

@inproceedings{hvarfner-iclr22,
  title        = {$\pi${BO}: Augmenting Acquisition Functions with User Beliefs for Bayesian Optimization},
  author       = {C. Hvarfner and D. Stoll and A. Souza and L. Nardi and M. Lindauer and F. Hutter},
  year         = 2022,
  booktitle    = {Proc. of {ICLR}'22}
}

@Manual{lopezibanez-github25,
  title = {iraceplot: Plots for Visualizing the Data Produced by the 'irace' Package},
  author = {M. López-Ibáñez and P. {Oñate Marín} and L. {Pérez Cáceres}},
  year = {2025},
  _note = {R package version 2.1.0, https://github.com/auto-optimization/iraceplot/},
  url = {https://auto-optimization.github.io/iraceplot/},
}

@inproceedings{wang-iui20ab,
    author = {Wang, D. and Ram, P. and Weidele, D. and Liu, S. and Muller, M. and Weisz, J. and Valente, A. and Chaudhary, A. and Torres, D. and Samulowitz, H. and Amini, L.},
    title = {AutoAI: Automating the End-to-End AI Lifecycle with Humans-in-the-Loop},
    pages = {77–78},
    booktitle = {Proceedings of the 25th International Conference on Intelligent User Interfaces ({IUI}'20)},
    year = {2020}
}

@inproceedings{biedenkapp-aaai17a,
  title        = {Efficient Parameter Importance Analysis via Ablation with Surrogates},
  author       = {A. Biedenkapp and M. Lindauer and K. Eggensperger and C. Fawcett and H. Hoos and F. Hutter},
  pages        = {773--779},
  crossref     = {aaai17},
}

@inproceedings{biedenkapp-lion18a,
  title        = {{CAVE}: Configuration Assessment, Visualization and Evaluation},
  author       = {A. Biedenkapp and J. Marben and M. Lindauer and F. Hutter},
  crossref     = {lion18},
}

@article{bischl-dmkd23a,
  title        = {Hyperparameter Optimization: Foundations, Algorithms, Best Practices, and Open Challenges},
  author       = {B. Bischl and M. Binder and M. Lang and T. Pielok and J. Richter and S. Coors and J. Thomas and T. Ullmann and M. Becker and A.{-}L. Boulesteix and D. Deng and M. Lindauer},
  year         = 2023,
  journal      = wileyirdmkd,
  publisher    = {Wiley Online Library},
  pages        = {e1484},
}

@inproceedings{drozdal-iui20a,
  title        = {Trust in {AutoML}: Exploring information needs for establishing trust in automated machine learning systems},
  author       = {J. Drozdal and J. Weisz and D. Wang and G. Dass and B. Yao and C. Zhao and M. J. Muller and L. Ju and H. Su},
  pages        = {297--307},
  crossref     = {iui20},
}

@article{elsken-arxiv22a,
  title        = {Neural architecture search for dense prediction tasks in computer vision},
  author       = {T. Elsken and A. Zela and J. Metzen and B. Staffler and T. Brox and A. Valada and F. Hutter},
  year         = 2022,
  journal      = {arXiv:2202.07242 [cs.CV]},
}

@article{feurer-jmlr22a,
  title        = {{Auto-Sklearn} 2.0: Hands-free AutoML via Meta-Learning},
  author       = {M. Feurer and K. Eggensperger and S. Falkner and M. Lindauer and F. Hutter},
  year         = 2022,
  journal      = jmlr,
  volume       = 23,
  number       = 261,
  pages        = {1--61},
  editor       = {M. Schoenauer},
}

@article{friedman-as01a,
  title        = {Greedy Function Approximation: A Gradient Boosting Machine},
  author       = {J. Friedman},
  year         = 2001,
  journal      = {Annals of Statistics},
  publisher    = {JSTOR},
  pages        = {1189--1232},
  volumne      = 29,
}

@inproceedings{golovin-kdd17a,
  title        = {Google {V}izier: A Service for Black-Box Optimization},
  author       = {D. Golovin and B. Solnik and S. Moitra and G. Kochanski and J. Karro and D. Sculley},
  pages        = {1487--1495},
  crossref     = {kdd17},
}

@inproceedings{hutter-icml14a,
  title        = {An Efficient Approach for Assessing Hyperparameter Importance},
  author       = {F. Hutter and H. Hoos and K. Leyton-Brown},
  pages        = {754--762},
  crossref     = {icml14},
}

@inproceedings{jin-sigkdd19a,
  title        = {{Auto-Keras}: An Efficient Neural Architecture Search System},
  author       = {H. Jin and Q. Song and X. Hu},
  pages        = {1946--1956},
  crossref     = {kdd19},
}

@article{lindauer-jmlr22a,
  title        = {{SMAC3}: A Versatile Bayesian Optimization Package for {H}yperparameter {O}ptimization},
  author       = {M. Lindauer and K. Eggensperger and M. Feurer and A. Biedenkapp and D. Deng and C. Benjamins and T. Ruhkopf and R. Sass and F. Hutter},
  year         = 2022,
  journal      = jmlr,
  volume       = 23,
  number       = 54,
  pages        = {1--9},
}

@inproceedings{mallik-neurips23a,
  title = {{PriorBand}: Practical Hyperparameter Optimization in the Age of Deep Learning},
  author = {N. Mallik and C. Hvarfner and E. Bergman and D. Stoll and M. Janowski   and M. Lindauer and L. Nardi and F. Hutter},
  crossref = {neurips23}
}

@inproceedings{moosbauer-neurips21a,
  title        = {Explaining Hyperparameter Optimization via Partial Dependence Plots},
  author       = {J. Moosbauer and J. Herbinger and G. Casalicchio and M. Lindauer and B. Bischl},
  crossref     = {neurips21},
  pages = {2280--2291}
}

@inproceedings{segel-automl23a,
  title        = {Symbolic Explanations for Hyperparameter Optimization},
  author       = {S. Segel and H. Graf and A. Tornede and B. Bischl and M. Lindauer},
  year         = 2023,
  crossref     = {automlconf23},
}

@inproceedings{thornton-kdd13a,
  title        = {{A}uto-{WEKA}: combined selection and {H}yperparameter {O}ptimization of classification algorithms},
  author       = {C. Thornton and F. Hutter and H. Hoos and K. Leyton-Brown},
  pages        = {847--855},
  crossref     = {kdd13},
}

@inproceedings{turner-neuripscomp21a,
  title        = {Bayesian Optimization is Superior to Random Search for Machine Learning Hyperparameter Tuning: Analysis of the {Black-Box Optimization Challenge 2020}},
  author       = {Turner, R. and Eriksson, D. and McCourt, M. and Kiili, J. and Laaksonen, E. and Xu, Z. and Guyon, I.},
  year         = 2021,
  pages        = {3--26},
  crossref     = {neuripscd20},
}

@article{zimmer-tpami21a,
  title        = {{Auto-Pytorch}: Multi-Fidelity MetaLearning for Efficient and Robust {AutoDL}},
  author       = {L. Zimmer and M. Lindauer and F. Hutter},
  year         = 2021,
  journal      = tpami,
  volume       = 43,
  pages        = {3079--3090},
  issue        = 9,
}

@proceedings{aaai17,
  title        = {Proceedings of the Thirty-First Conference on Artificial Intelligence ({AAAI}'17)},
  year         = 2017,
  booktitle    = {Proceedings of the Thirty-First Conference on Artificial Intelligence ({AAAI}'17)},
  publisher    = {{AAAI} Press},
  editor       = {S. Singh and S. Markovitch},
}

@proceedings{automl18,
  title        = {{ICML} workshop on Automated Machine Learning ({Auto{ML}} workshop 2018)},
  year         = 2018,
  booktitle    = {{ICML} workshop on Automated Machine Learning ({Auto{ML}} workshop 2018)},
  editor       = {R. Garnett and F. Hutter and J. Vanschoren and P. Brazdil and R. Caruana and C. Giraud-Carrier and I. Guyon and B. Kégl},
}

@proceedings{automlconf23,
  title        = {Proceedings of the Second International Conference on Automated Machine Learning},
  year         = 2023,
  booktitle    = {Proceedings of the Second International Conference on Automated Machine Learning},
  publisher    = {Proceedings of Machine Learning Research},
  editor       = {A. Faust and C. White and F. Hutter and R. Garnett and J. Gardner},
}

@proceedings{icml14,
  title        = {Proceedings of the 31th International Conference on Machine Learning, ({ICML}'14)},
  year         = 2014,
  booktitle    = {Proceedings of the 31th International Conference on Machine Learning, ({ICML}'14)},
  publisher    = {Omnipress},
  editor       = {E. Xing and T. Jebara},
}

@STRING{acm     = "ACM Press" }

@STRING{aaai    = "Proceedings of the National Conference on Artificial
                  Intelligence (AAAI)" }

@STRING{ai      = "Artificial Intelligence" }

@STRING{ieee = "IEEE" }

@STRING{is      = "Informatik-Spektrum" }

@STRING{jmlr    = "Journal of Machine Learning Research" }

@STRING{springer = "Springer" }

@STRING{tpami =   "IEEE Transactions on Pattern Analysis and Machine Intelligence"}

@STRING{wiley   = "John Wiley \& sons" }

@STRING{iui     = "Proceedings of the Conference on Intelligent User Interfaces"}

@STRING{wileyirdmkd   = "Wiley Interdisciplinary Reviews: Data Mining and Knowledge Discovery"}

\end{document}